# Inverse Reinforcement Learning with Simultaneous Estimation of Rewards and Dynamics

Michael Herman\*[†]     Tobias Gindele\*     Jörg Wagner\*     Felix Schmitt\*     Wolfram Burgard[†]

\*Robert Bosch GmbH
D-70442 Stuttgart, Germany

[†]University of Freiburg
D-79110 Freiburg, Germany

## Abstract

Inverse Reinforcement Learning (IRL) describes the problem of learning an unknown reward function of a Markov Decision Process (MDP) from observed behavior of an agent. Since the agent's behavior originates in its policy and MDP policies depend on both the stochastic system dynamics as well as the reward function, the solution of the inverse problem is significantly influenced by both. Current IRL approaches assume that if the transition model is unknown, additional samples from the system's dynamics are accessible, or the observed behavior provides enough samples of the system's dynamics to solve the inverse problem accurately. These assumptions are often not satisfied. To overcome this, we present a gradient-based IRL approach that simultaneously estimates the system's dynamics. By solving the combined optimization problem, our approach takes into account the bias of the demonstrations, which stems from the generating policy. The evaluation on a synthetic MDP and a transfer learning task shows improvements regarding the sample efficiency as well as the accuracy of the estimated reward functions and transition models.

## 1 Introduction

With more and more autonomous systems performing complex tasks in various applications, it is necessary to provide simple programming approaches for non-experts to adjust the systems' abilities to new environments. A mathematical framework for modeling decision making under partly random outcomes are MDPs. By specifying an environment, its dynamics and a reward function, optimal policies can be derived, e.g. by reinforcement learning (RL) (Sutton and Barto, 1998). However, when the environment or the problem gets complex, it is often difficult to specify appropriate reward functions that yield a desired behavior. Instead, it can be easier to provide demonstrations of the desired behavior. Therefore, Ng and Russell (2000) introduced Inverse Reinforcement Learning (IRL), which describes the problem of recovering a reward function of an MDP from demonstrations.

The basic idea of IRL is that the reward function is the most succinct representation of an expert's objective, which can be easily transferred to new environments. Many approaches have been proposed to solve the IRL problem, such as (Abbeel and Ng, 2004; Ratliff et al., 2006; Neu and Szepesvári, 2007; Ramachandran and Amir, 2007; Rothkopf and Dimitrakakis, 2011). Since expert demonstrations are rarely optimal, IRL approaches have been introduced that deal with stochastic behavior, e.g. (Ziebart et al., 2008, 2010; Bloem and Bambos, 2014). Most of these approaches have in common that they require the system's dynamics to be known. Since this assumption is often not satisfied, model-free IRL algorithms have been proposed, such as (Boularias et al., 2011; Tossou and Dimitrakakis, 2013; Klein et al., 2012, 2013).

As the model-based approaches need to repeatedly solve the RL problem as part of solving the IRL problem, they require an accurate model of the system's dynamics. Most of them assume that an MDP model including the dynamics is either given or can be estimated well enough from demonstrations. However, as the observations are the result of an expert's policy, they only provide demonstrations of desired behavior in desired states. As a consequence, it is often not possible to estimate an accurate transition model directly from expert demonstrations. Model-free approaches typically require that the observed demon-strations contain enough samples of the system's dynamics to accurately learn the reward function or require access to the environment or a simulator to generate additional data. However, in many applications realistic simulators do not exist and it is not possible to query the environment. In this case, current model-free IRL approaches either don't consider rewards or transitions of unobserved states and actions, or tend to suffer from wrong generalizations due to heuristics.

We argue that simultaneously optimizing the likelihood of the demonstrations with respect to the reward function and the dynamics of the MDP can improve the accuracy of the estimates and with it the resulting policy. Even though many transitions have never been observed, they can to some degree be inferred by taking into account that the data has been generated by an expert's policy. Since the expert's policy is the result of both his reward function and his belief about the system's dynamics, the frequency of state-action pairs in the data carries information about the expert's objective. This can be exploited to improve the sample efficiency and the accuracy of the estimation of the system's dynamics and the reward function, as they both influence the policy. One side of this bilateral influence has been used by Golub et al. (2013), who showed that more accurate dynamics can be estimated when the reward function is known.

Our contribution is integrating the learning of the transition model into the IRL framework, by considering that demonstrations have been generated based on a policy. This even allows drawing conclusions about parts of the transition model that were never observed. We provide a general gradient-based solution for a simultaneous estimation of rewards and dynamics (SERD). Furthermore, we derive a concrete algorithm, based on Maximum Discounted Causal Entropy IRL (Bloem and Bambos, 2014). Part of it is an iterative computation of the policy gradient for which we show convergence. We evaluate our approach on synthetic MDPs, compare it to model-based and model-free IRL approaches, and test its generalization capabilities in a transfer task. More detailed derivations and proofs are provided in the supplementary material.

## 2 Fundamentals

This section introduces the notation and fundamentals to formulate the IRL problem with a simultaneous estimation of rewards and dynamics.

### 2.1 Markov Decision Processes

An MDP is a tuple $M = \{S, A, P(s'|s,a), \gamma, R, P(s_0)\}$, where $S$ is the state space with states $s \in S$, $A$ is the action space with actions $a \in A$, $P(s'|s,a)$ is the probability of a transition to $s'$ when action $a$ is applied in state $s$, $\gamma \in [0,1)$ is a discount factor, $R : S \times A \to \mathbb{R}$ is a reward function which assigns a real-valued reward for picking action $a$ in state $s$, and $P(s_0)$ is a start state probability distribution.

The goal of an MDP is to find an optimal policy $\pi^*(s,a) = P(a|s)$, which specifies the probability of taking action $a$ in state $s$, such that executing this policy maximizes the expected, discounted future reward:

$$V^\pi(s) = \mathbb{E}\left[\sum_{t=0}^{\infty} \gamma^t R(s_t, a_t) \,|\, s_0 = s, \pi\right] \quad (1)$$

If the policy is deterministic, the probability distribution can be expressed with one single action value $\pi_d^*(s) = a$. Additionally, a Q-function can be defined, which specifies the expected, discounted, cumulated reward for starting in state $s$, picking action $a$ and then following the policy $\pi$.

$$Q^\pi(s,a) = \mathbb{E}\left[\sum_{t=0}^{\infty} \gamma^t R(s_t, a_t) \,|\, s_0 = s, a_0 = a, \pi\right] \quad (2)$$

### 2.2 Inverse Reinforcement Learning

IRL describes the problem of learning the unknown reward function of an MDP from observed behavior of an agent acting according to some stochastic policy. It is therefore characterized by the tuple $M \setminus R$ and observed demonstrations $D = \{\tau_1, \tau_2, \ldots, \tau_N\}$ with trajectories $\tau = \{(s_0^\tau, a_0^\tau), (s_1^\tau, a_1^\tau), \ldots, (s_{T_\tau}^\tau, a_{T_\tau}^\tau)\}$. The goal of IRL is to estimate the agent's reward function $R(s,a)$, which explains the observed behavior in the demonstrations. Often this reward is expressed as state- and action-dependent features $\boldsymbol{f} : S \times A \to \mathbb{R}^d$. An IRL model assuming stochastic expert behavior is the Maximum Entropy IRL (MaxEnt IRL) model of Ziebart et al. (2008), which has been applied to different learning problems, such as (Ziebart et al., 2009; Henry et al., 2010; Kuderer et al., 2013). Since it doesn't support stochastic transition models to the full extent, Ziebart et al. (2010) proposed an approach, called Maximum Causal Entropy IRL (MCE IRL). Bloem and Bambos (2014) extended MCE IRL to the infinite time horizon case, which is called Maximum Discounted Causal Entropy IRL (MDCE IRL). They derive a simplified stationary soft value iteration solution for MDCE IRL, which is formulated as

$$V_{\boldsymbol{\theta}}(s) = \log\left(\sum_{a \in A} \exp(Q_{\boldsymbol{\theta}}(s,a))\right) \quad (3)$$

Michael Herman, Tobias Gindele, Jörg Wagner, Felix Schmitt, Wolfram Burgardwhere the soft state-action value $Q_\theta(s,a)$ is defined as

$$Q_{\boldsymbol{\theta}}(s,a) = \boldsymbol{\theta}^T \boldsymbol{f}(s,a) + \gamma \sum_{s' \in S} P(s'|s,a) V_{\boldsymbol{\theta}}(s'). \quad (4)$$

This soft value iteration is a contraction mapping, which has been proven in (Bloem and Bambos, 2014). The stochastic policy $\pi_\theta(s,a)$ can be extracted from the stationary fixed point solutions of $V_{\boldsymbol{\theta}}(s)$ and $Q_{\boldsymbol{\theta}}(s,a)$ and forms a Boltzmann distribution over the Q-values of all valid actions in state $s$:

$$\begin{aligned}\pi_{\boldsymbol{\theta}}(s,a) &= \exp(Q_{\boldsymbol{\theta}}(s,a) - V_{\boldsymbol{\theta}}(s))\\ &= \frac{\exp(Q_{\boldsymbol{\theta}}(s,a))}{\sum_{a' \in A} \exp(Q_{\boldsymbol{\theta}}(s,a'))}.\end{aligned} \quad (5)$$

The linear combination of feature weights $\boldsymbol{\theta} \in \mathbb{R}^d$ and features $\boldsymbol{f}(s,a) \in \mathbb{R}^d$ in Eq. (4) can be interpreted as a reward function $R(s,a) = \sum_{k=1}^{d} \theta_k f_k(s,a)$. As the features are defined by the states and actions, estimating a reward function degrades to finding appropriate feature weights. Due to the Markov assumption in MDPs it is possible to formulate the probability of a specific trajectory $\tau$ based on the start distribution, the single actions and the transitions:

$$P(\tau|M,\boldsymbol{\theta}) = P(s_0^\tau) \prod_{t=0}^{T_\tau - 1} \left[ \pi_{\boldsymbol{\theta}}(s_t^\tau, a_t^\tau) \right. \\ \left. \cdot P(s_{t+1}^\tau | s_t^\tau, a_t^\tau) \right]. \quad (6)$$

Ziebart et al. have shown in (Ziebart, 2010; Ziebart et al., 2010) that appropriate feature weights can be learned by optimizing the likelihood of the data under the maximum causal entropy distribution policy $\pi_\theta(s,a)$ from Eq. (5). Assuming independent trajectories, the likelihood of the demonstrations in $D$ can be expressed as

$$P(D|M,\boldsymbol{\theta}) = \prod_{\tau \in D} P(\tau|M,\boldsymbol{\theta}). \quad (7)$$

Meaningful feature weights can then be found by maximizing the log-likelihood of the demonstrations with respect to the feature weights $\boldsymbol{\theta}$:

$$\boldsymbol{\theta}^* = \operatorname*{argmax}_{\boldsymbol{\theta}} \log P(D|M,\boldsymbol{\theta}). \quad (8)$$

## 3 Simultaneous Estimation of Rewards and Dynamics (SERD)

Many IRL approaches assume that the system dynamics are known or can be estimated well enough from demonstrations. To the best of our knowledge, the robustness of IRL against wrong transition models has not been studied so far, even though this problem has already been pointed out in (Ramachandran and Amir, 2007). However, as the transition model influences the policy, the reward estimation of the IRL problem can be falsified due to wrong transition model estimates. It may then be advantageous to learn both at once, in order to capture the relationship between the reward function and the dynamics model in the policy. Additionally, it is possible that the agent's belief about the system's dynamics differs from the true one, which yields a wrong policy. This led us to the formulation of a new problem class, which can be characterized as follows:

**Determine:**

- Agent's reward function $R(s,a)$
- Agent's belief about the dynamics $P_A(s'|s,a)$
- Real dynamics $P(s'|s,a)$

**Given**

- MDP $M \setminus \{R, P(s'|s,a), P_A(s'|s,a)\}$ without the reward function or any dynamics
- Demonstrations $D$ of an agent acting in $M$ based on a policy that depends on $R(s,a)$ and $P_A(s'|s,a)$

To solve this problem we propose an approach with a combined estimation, called Simultaneous Estimation of Rewards and Dynamics (SERD). We assume that there exist models for $P(s'|s,a)$ and $P_A(s'|s,a)$, which can be parameterized. Therefore, we introduce a set of parameters, which should be estimated from the given demonstrations $D$:

$\boldsymbol{\theta}_R$    Feature weights of the reward function $R(s,a)$
$\boldsymbol{\theta}_{T_A}$    Parameters of the agent's transition model $P_{\boldsymbol{\theta}_{T_A}}$
$\boldsymbol{\theta}_T$    Parameters of the real transition model $P_{\boldsymbol{\theta}_T}$

Since no prior information about rewards or dynamics is known, our SERD approach for solving the problem is to maximize the likelihood of the demonstrations with respect to these parameters, which can be combined in the parameter vector $\boldsymbol{\theta} = \begin{pmatrix} \boldsymbol{\theta}_R^\mathsf{T} & \boldsymbol{\theta}_{T_A}^\mathsf{T} & \boldsymbol{\theta}_T^\mathsf{T} \end{pmatrix}^\mathsf{T}$. This is related to the approaches in (Ziebart, 2010; Ziebart et al., 2008), which estimates feature weights $\boldsymbol{\theta}_R$ by maximizing the log likelihood of the demonstrations under the maximum entropy distribution policy $\pi_\theta(s,a)$ of Eq. (5). Assuming independent trajectories, the likelihood of the demonstrations in $D$ can be expressed as

$$P(D|M,\boldsymbol{\theta}) = \prod_{\tau \in D} P(s_0^\tau) \prod_{t=0}^{T_\tau - 1} \left[ \pi_{\boldsymbol{\theta}}(s_t^\tau, a_t^\tau) \right. \\ \left. \cdot P_{\boldsymbol{\theta}_T}(s_{t+1}^\tau | s_t^\tau, a_t^\tau) \right]. \quad (9)$$

Inverse Reinforcement Learning with Simultaneous Estimation of Rewards and Dynamics

We want to point out that the policy $\pi_{\boldsymbol{\theta}}(s,a)$ depends on both feature weights $\boldsymbol{\theta}_R$ as well as the agent's dynamics parameters $\boldsymbol{\theta}_{T_A}$, whereas the transition model $P_{\boldsymbol{\theta}_T}(s'|s,a)$ only depends on $\boldsymbol{\theta}_T$. The log likelihood of the demonstrations $L_{\boldsymbol{\theta}}(D)$ can then be derived from Eq. (9):

$$L_{\boldsymbol{\theta}}(D) = \log P(D|M, \boldsymbol{\theta}) \tag{10}$$

$$= \sum_{\tau \in D} \left[ \log P(s_0^\tau) + \sum_{t=0}^{T_\tau - 1} \left[ \log \pi_{\boldsymbol{\theta}}(s_t^\tau, a_t^\tau) \right. \right.$$
$$\left. \left. + \log P_{\boldsymbol{\theta}_T}(s_{t+1}^\tau | s_t^\tau, a_t^\tau) \right] \right]. \tag{11}$$

Solving the SERD problem then corresponds to optimizing the log likelihood of the demonstrations with respect to $\boldsymbol{\theta}$, which is formulated as:

$$\boldsymbol{\theta}^* = \operatorname*{argmax}_{\boldsymbol{\theta}} L_{\boldsymbol{\theta}}(D). \tag{12}$$

We propose a gradient-based method to optimize the log likelihood of Eq. (11), which shares similarities with the approach in (Neu and Szepesvári, 2007). Therefore, we derive the gradient $\nabla_{\boldsymbol{\theta}} L_{\boldsymbol{\theta}}(D)$ to solve the SERD problem of Eq. (12). This gradient can be formalized as:

$$\nabla_{\boldsymbol{\theta}} L_{\boldsymbol{\theta}}(D) = \sum_{\tau \in D} \sum_{t=0}^{T_\tau - 1} \left[ \nabla_{\boldsymbol{\theta}} \log \pi_{\boldsymbol{\theta}}(s_t^\tau, a_t^\tau) \right.$$
$$\left. + \nabla_{\boldsymbol{\theta}} \log P_{\boldsymbol{\theta}_T}(s_{t+1}^\tau | s_t^\tau, a_t^\tau) \right]. \tag{13}$$

As the start state distribution does not depend on any of the parameters $\boldsymbol{\theta}$ the term $\sum_{\tau \in D} \log P(s_0^\tau)$ vanishes. The gradient $\nabla_{\boldsymbol{\theta}} L_{\boldsymbol{\theta}}(D)$ can therefore be factorized to cumulated gradients of the state-action pair probability $\nabla_{\boldsymbol{\theta}} \log \pi_{\boldsymbol{\theta}}(s_t^\tau, a_t^\tau)$ and the transition probability $\nabla_{\boldsymbol{\theta}} \log P_{\boldsymbol{\theta}_T}(s_{t+1}^\tau | s_t^\tau, a_t^\tau)$. Since the gradient of the true transition probability is model dependent, the following derivations will focus on the gradient of the policy $\nabla_{\boldsymbol{\theta}} \log \pi_{\boldsymbol{\theta}}(s_t^\tau, a_t^\tau)$. Usually, this policy is problem specific, which requires to specify an IRL type. We will derive the gradient for MDCE IRL, but an extension to further IRL solutions and policies is possible.

## 4 Maximum Discounted Causal Entropy SERD (MDCE-SERD)

In the following, we will exemplarily derive the gradient of policies $\nabla_{\boldsymbol{\theta}} \log \pi_{\boldsymbol{\theta}}(s_t^\tau, a_t^\tau)$ that are based on MDCE IRL by Bloem and Bambos (2014), which has been introduced in Section 2.2. Under this assumption the partial derivative of the policy can be decomposed by replacing $\pi_{\boldsymbol{\theta}}(s_t^\tau, a_t^\tau)$ through its representation in MDCE IRL:

$$\frac{\partial}{\partial \theta_i} \log \pi_{\boldsymbol{\theta}}(s,a) = \frac{\partial}{\partial \theta_i} Q_{\boldsymbol{\theta}}(s,a)$$
$$- \mathbb{E}_{\pi_{\boldsymbol{\theta}}(s,a')} \left[ \frac{\partial}{\partial \theta_i} Q_{\boldsymbol{\theta}}(s,a') \right]. \tag{14}$$

It follows that the gradient of the policy depends on the gradient of the state-action value function $\frac{\partial}{\partial \theta_i} Q_{\boldsymbol{\theta}}(s,a)$ and the expected gradient $\mathbb{E}_{\pi_{\boldsymbol{\theta}}(s,a')} \left[ \frac{\partial}{\partial \theta_i} Q_{\boldsymbol{\theta}}(s,a') \right]$. As the expectation is taken with respect to the stochastic policy $\pi_{\boldsymbol{\theta}}(s,a')$, which depends on the Q-function, a basic requirement for its computation is a converged Q- and value-function. In the following, we will provide the partial derivative of the iterative soft Q-function with respect to $\theta_i$, which we call soft Q-gradient. The partial derivatives with respect to the individual parameter types, such as feature weights or transition parameters, can be found in the supplement.

$$\frac{\partial}{\partial \theta_i} Q_{\boldsymbol{\theta}}(s,a) = \frac{\partial}{\partial \theta_i} \boldsymbol{\theta}_R^\mathsf{T} \boldsymbol{f}(s,a) \tag{15}$$

$$+ \gamma \sum_{s' \in S} \left[ \left( \frac{\partial}{\partial \theta_i} P_{\boldsymbol{\theta}_{T_A}}(s'|s,a) \right) V_{\boldsymbol{\theta}}(s') \right] \tag{16}$$

$$+ \gamma \sum_{s' \in S} \left\{ P_{\boldsymbol{\theta}_{T_A}}(s'|s,a) \right.$$
$$\left. \cdot \mathbb{E}_{\pi_{\boldsymbol{\theta}}(s',a')} \left[ \frac{\partial}{\partial \theta_i} Q_{\boldsymbol{\theta}}(s',a') \right] \right\} \tag{17}$$

The soft Q-gradient computation is a linear equation system and can thus be computed directly. Nevertheless, if the number of parameters is large, it can be beneficial to choose an iterative approach. For this purpose, Eq. (17) can be interpreted as a recursive function, which we call soft Q-gradient iteration. The first two terms (15) and (16) are constants due to the requirement of a static and converged soft Q-function. The third term (17) propagates expected gradients through the space of states $S$, actions $A$, and parameter dimensions $\boldsymbol{\theta}$. The partial derivative $\frac{\partial}{\partial \theta_i} Q_{\boldsymbol{\theta}}(s,a)$ is a fixed point iteration and can therefore be computed by starting with arbitrary gradients and recursively applying Eq. (17). We will prove this in section 4.1.

Solving the linear system of Eq. (17) with $|S|$ states, $|A|$ actions, and $N_{\boldsymbol{\theta}}$ parameters directly via LU decompositions requires $\mathcal{O}\left(N_{\boldsymbol{\theta}} \cdot (|S| \cdot |A|)^3\right)$ computations. Instead, a single iteration of Eq. (17) requires $\mathcal{O}\left(N_{\boldsymbol{\theta}} \cdot (|S| \cdot |A|)^2\right)$ computations, which can be beneficial if the number of states and actions is large. Additionally, the result of the soft Q-iteration can be used to initialize a subsequent soft Q-iteration with slightly changed parameters to further decrease the number of necessary iterations.

Michael Herman, Tobias Gindele, Jörg Wagner, Felix Schmitt, Wolfram Burgard

Algorithm 1 summarizes the MDCE-SERD algorithm. The function *DynamicsEstimator* provides a naive dynamics estimate from the transitions of the demonstrations. *SoftQIteration* performs the soft Q-iteration from Eq. (3) and (4) until convergence. *DerivePolicy* performs a policy update based on the policy definition in Eq. (5) and the function *SoftQGradientIteration* solves the soft Q-gradient iteration from Eq. (17) until convergence. Then, the function *ComputeGradient* calculates the gradient based on Eq. (13).

**Algorithm 1** MDCE-SERD algorithm

**Require:** MDP $M \setminus \{R, P_T, P_{T_A}\}$, Demonstrations $D$, initial $\tilde{\boldsymbol{\theta}}$, step size $\alpha : \mathbb{N}_+ \to \mathbb{R}_+$, $t \leftarrow 0$
$\boldsymbol{\theta}_0 \leftarrow \text{DynamicsEstimator}(M, D, \tilde{\boldsymbol{\theta}})$
**while** not sufficiently converged **do**
    $\boldsymbol{Q}_{\boldsymbol{\theta}} \leftarrow \text{SoftQIteration}(M, \boldsymbol{\theta}_t)$
    $\pi_{\boldsymbol{\theta}} \leftarrow \text{DerivePolicy}(M, \boldsymbol{Q}_{\boldsymbol{\theta}})$
    $d\boldsymbol{Q}_{\boldsymbol{\theta}} \leftarrow \text{SoftQGradientIteration}(M, \boldsymbol{Q}_{\boldsymbol{\theta}}, \pi_{\boldsymbol{\theta}}, \boldsymbol{\theta}_t)$
    $\nabla_{\boldsymbol{\theta}} L_{\boldsymbol{\theta}}(D) \leftarrow \text{ComputeGradient}(M, D, d\boldsymbol{Q}_{\boldsymbol{\theta}})$
    $\boldsymbol{\theta}_{t+1} \leftarrow \boldsymbol{\theta}_t + \alpha(t)\nabla_{\boldsymbol{\theta}} L_{\boldsymbol{\theta}}(D)$
    $t \leftarrow t + 1$
**end while**

### 4.1 Proofs

To prove the correctness and convergence of the proposed algorithm, it must be shown that the soft Q-iteration is a contraction mapping, that the soft Q-function is differentiable, and that the soft Q-gradient iteration is a contraction mapping. Bloem and Bambos (2014) have shown that the soft value iteration operator is a contraction mapping. The proof for the soft Q-iteration is straightforward and is presented in the supplementary material together with more detailed derivations for all proofs.

**Theorem 4.1.** *The soft Q-iteration operator $T_{\boldsymbol{\theta}}^{soft}(\boldsymbol{Q})$ is a contraction mapping with only one fixed point. Therefore, it is Lipschitz continuous $||T_{\boldsymbol{\theta}}^{soft}(\boldsymbol{Q}_m) - T_{\boldsymbol{\theta}}^{soft}(\boldsymbol{Q}_n)||_\infty \leq L||\boldsymbol{Q}_m - \boldsymbol{Q}_n||_\infty$ for all $\boldsymbol{Q}_m, \boldsymbol{Q}_n \in \mathbb{R}^{|S| \times |A|}$ with a Lipschitz constant $L = \gamma \in [0, 1)$.*

**Theorem 4.2.** *The converged soft Q-function is differentiable with respect to $\boldsymbol{\theta}$.*

The soft Q-gradient operator $U_{\boldsymbol{\theta}}^{soft}(\boldsymbol{\Phi}) \in \mathbb{R}^{|S| \times |A| \times |\Psi|}$ is defined as:

$$U_{\boldsymbol{\theta}}^{soft}(\boldsymbol{\Phi})(s,a,i) = \frac{\partial}{\partial \theta_i} \boldsymbol{\theta}_R^\intercal \boldsymbol{f}(s,a)$$
$$+ \gamma \sum_{s' \in S} \left[ \left( \frac{\partial}{\partial \theta_i} P_{\boldsymbol{\theta}_{T_A}}(s'|s,a) \right) V_{\boldsymbol{\theta}}(s') \right]$$
$$+ \gamma \sum_{s' \in S} \left\{ P_{\boldsymbol{\theta}_{T_A}}(s'|s,a) \cdot \mathbb{E}_{\pi_{\boldsymbol{\theta}}(s',a')} [\Phi(s',a',i)] \right\},$$

for all $s \in S, a \in A$ and parameter dimensions $i \in \Psi$ with $\Psi = \{1, \ldots, \dim(\boldsymbol{\theta})\}$ and the gradient

$$\Phi(s,a,i) = \frac{\partial}{\partial \theta_i} Q_{\boldsymbol{\theta}}(s,a).$$

Some auxiliary lemmata and definitions are necessary to prove that the Q-gradient iteration is a contraction mapping. In order to argue about the monotonicity of multidimensional functions, a partial order on $\mathbb{R}^{A \times B \times C}$ is introduced. The monotonicity of the operator $U_{\boldsymbol{\theta}}^{soft}(\boldsymbol{\Phi})$ with respect to the introduced partial order is proven.

**Definition 4.3.** *For $\boldsymbol{x}, \boldsymbol{y} \in \mathbb{R}^{A \times B \times C}$ with $A, B, C \in \mathbb{N}^+$, the partial order $\preceq$ is defined as $\boldsymbol{x} \preceq \boldsymbol{y} \Leftrightarrow \forall a \in A, b \in B, c \in C : x_{a,b,c} \leq y_{a,b,c}$.*

**Lemma 4.4.** *The soft Q-gradient iteration operator $U_{\boldsymbol{\theta}}^{soft}(\boldsymbol{\Phi})(s,a,i)$ is monotone, satisfying $\forall \boldsymbol{\Phi}_m, \boldsymbol{\Phi}_n \in \mathbb{R}^{|S| \times |A| \times |\Psi|} : \boldsymbol{\Phi}_m \preceq \boldsymbol{\Phi}_n \to U_{\boldsymbol{\theta}}^{soft}(\boldsymbol{\Phi}_m) \preceq U_{\boldsymbol{\theta}}^{soft}(\boldsymbol{\Phi}_n).*$

*Proof.* The partial derivative of the $U_{\boldsymbol{\theta}}^{soft}(\boldsymbol{\Phi})(s,a,i)$ with respect to a single value $\Phi(s_k, a_k, k)$ is

$$\frac{\partial}{\partial \Phi(s_k, a_k, k)} U_{\boldsymbol{\theta}}^{soft}(\boldsymbol{\Phi})(s,a,i)$$
$$= \frac{\partial}{\partial \Phi(s_k, a_k, k)} \gamma \sum_{s' \in S} \Big\{ P_{\boldsymbol{\theta}_{T_A}}(s'|s,a)$$
$$\cdot \mathbb{E}_{\pi_{\boldsymbol{\theta}}(s',a')} [\Phi(s',a',i)] \Big\}$$
$$= \gamma P_{\boldsymbol{\theta}_{T_A}}(s_k|s,a) \pi_{\boldsymbol{\theta}}(s_k, a_k).$$

From the definition of the MDP it follows that $\gamma \in [0,1)$ and the probability distributions $P_{\boldsymbol{\theta}_{T_A}}(s_i|s,a) \in [0,1]$ as well as $\pi_{\boldsymbol{\theta}}(s_k, a_k) \in [0,1]$. Since all terms of the partial derivative $\frac{\partial}{\partial \Phi(s_k, a_k, k)} U_{\boldsymbol{\theta}}^{soft}(\boldsymbol{\Phi})(s,a,i)$ are positive or zero, it follows that $\frac{\partial}{\partial \Phi(s_k, a_k, k)} U_{\boldsymbol{\theta}}^{soft}(\boldsymbol{\Phi})(s,a,i) \geq 0$. □

**Theorem 4.5.** *The soft Q-gradient iteration operator $U_{\boldsymbol{\theta}}^{soft}(\boldsymbol{\Phi})(s,a,i)$ is a contraction mapping with only one fixed point. Therefore, it is Lipschitz continuous $||U_{\boldsymbol{\theta}}^{soft}(\boldsymbol{\Phi}_m) - U_{\boldsymbol{\theta}}^{soft}(\boldsymbol{\Phi}_n)||_\infty \leq L||\boldsymbol{\Phi}_m - \boldsymbol{\Phi}_n||_\infty$ for all $\boldsymbol{\Phi}_m, \boldsymbol{\Phi}_n \in \mathbb{R}^{|S| \times |A| \times |\Psi|}$ with a Lipschitz constant $L \in [0,1)$.*

*Proof.* Consider $\boldsymbol{\Phi}_m, \boldsymbol{\Phi}_n \in \mathbb{R}^{|S| \times |A| \times |\Psi|}$. There exists a distance $d$ under the supremum norm, for which $\exists d \in \mathbb{R}_0^+ : ||\boldsymbol{\Phi}_m - \boldsymbol{\Phi}_n||_\infty = d$ holds and therefore $-d\boldsymbol{1} \preceq \boldsymbol{\Phi}_m - \boldsymbol{\Phi}_n \preceq -d\boldsymbol{1}$ with $\boldsymbol{1} = (1)_{k,l,m}$, where $1 \leq k \leq |S|, 1 \leq l \leq |A|, 1 \leq m \leq |\Psi|$. By adding $d$ to every element of $\boldsymbol{\Phi}_n$ it is guaranteed that $\boldsymbol{\Phi}_m \preceq \boldsymbol{\Phi}_n + d\boldsymbol{1}$. Therefore, the monotonicity condition of Lemma 4.4 is satisfied: $U_{\boldsymbol{\theta}}^{soft}(\boldsymbol{\Phi}_m) \preceq U_{\boldsymbol{\theta}}^{soft}(\boldsymbol{\Phi}_n + d\boldsymbol{1})$. Then, it follows $\forall s \in S, a \in A, i \in \Psi$:

$$
\begin{aligned}
&U_{\boldsymbol{\theta}}^{soft}(\boldsymbol{\Phi}_m)(s,a,i) \\
&\leq U_{\boldsymbol{\theta}}^{soft}(\boldsymbol{\Phi}_n + d\mathbf{1})(s,a,i) \\
&= \frac{\partial}{\partial \theta_i} \boldsymbol{\theta}_R^\intercal \boldsymbol{f}(s,a) + \gamma \sum_{s' \in S} \left[ \left( \frac{\partial}{\partial \theta_i} P_{\boldsymbol{\theta}_{T_A}}(s'|s,a) \right) V_{\boldsymbol{\theta}}(s') \right] \\
&\quad + \gamma \sum_{s' \in S} \left\{ P_{\boldsymbol{\theta}_{T_A}}(s'|s,a) \, \mathbb{E}_{\pi_{\boldsymbol{\theta}}(s',a')} [\Phi(s',a',i) + d] \right\} \\
&= \frac{\partial}{\partial \theta_i} \boldsymbol{\theta}_R^\intercal \boldsymbol{f}(s,a) + \gamma \sum_{s' \in S} \left[ \left( \frac{\partial}{\partial \theta_i} P_{\boldsymbol{\theta}_{T_A}}(s'|s,a) \right) V_{\boldsymbol{\theta}}(s') \right] \\
&\quad + \gamma \sum_{s' \in S} \left\{ P_{\boldsymbol{\theta}_{T_A}}(s'|s,a) \, \mathbb{E}_{\pi_{\boldsymbol{\theta}}(s',a')} [\Phi(s',a',i)] \right\} + \gamma d \\
&= U_{\boldsymbol{\theta}}^{soft}(\boldsymbol{\Phi}_n)(s,a,i) + \gamma d
\end{aligned}
$$

In vector notation this results in $U_{\boldsymbol{\theta}}^{soft}(\boldsymbol{\Phi}_m) \preceq U_{\boldsymbol{\theta}}^{soft}(\boldsymbol{\Phi}_n) + \gamma d\mathbf{1}$. From the symmetric definition of $d$ it equally follows that $\boldsymbol{\Phi}_n \preceq \boldsymbol{\Phi}_m + d$ and consequently $U_{\boldsymbol{\theta}}^{soft}(\boldsymbol{\Phi}_n) \preceq U_{\boldsymbol{\theta}}^{soft}(\boldsymbol{\Phi}_m) + \gamma d\mathbf{1}$. By combining these inequations, the Lipschitz continuity of the soft Q-gradient iteration can be shown:

$$
\begin{aligned}
\gamma d\mathbf{1} \preceq \quad & U_{\boldsymbol{\theta}}^{soft}(\boldsymbol{\Phi}_m) - U_{\boldsymbol{\theta}}^{soft}(\boldsymbol{\Phi}_n) \quad \preceq \gamma d\mathbf{1} \\
& ||U_{\boldsymbol{\theta}}^{soft}(\boldsymbol{\Phi}_m) - U_{\boldsymbol{\theta}}^{soft}(\boldsymbol{\Phi}_n)||_\infty \leq \gamma d \\
& ||U_{\boldsymbol{\theta}}^{soft}(\boldsymbol{\Phi}_m) - U_{\boldsymbol{\theta}}^{soft}(\boldsymbol{\Phi}_n)||_\infty \leq \gamma ||\boldsymbol{\Phi}_m - \boldsymbol{\Phi}_n||_\infty
\end{aligned}
$$

This proves that the soft Q-gradient iteration operator $U_{\boldsymbol{\theta}}^{soft}(\boldsymbol{\Phi})(s,a,i)$ is Lipschitz continuous with a Lipschitz constant $L = \gamma$ and $\gamma \in [0,1)$, resulting in a contraction mapping. As this holds for the whole input space $\mathbb{R}^{|S| \times |A| \times |\Psi|}$, two points would always contract, so there cannot exist two fixed points. □

## 5 Related Work

In prior work various model-free and model-based IRL approaches have been proposed. Ziebart et al. (2008) proposed MaxEnt IRL in an early work, where a maximum entropy probability distribution of trajectories is trained to match feature expectations. One drawback of MaxEnt IRL is that it does not account for stochastic transition models to the full extent. Therefore, Ziebart et al. (2010) extended the previous approach to MCE IRL, which allows for stochastic transition models. Both algorithms require the transition model of the MDP to be known and are computationally expensive, since they need to repeatedly solve the RL problem. Therefore, model-free IRL approaches have been proposed, which overcome this requirement. Boularias et al. (2011) propose an approach called Relative Entropy IRL (REIRL), which minimizes the Kullback-Leibler divergence between a learned trajectory distribution and one that is based on a baseline policy under the constraint to match feature expectations. For this purpose, the baseline policy is approximated via importance sampling from arbitrary policies. Since the problem formulation of REIRL originates from MaxEnt IRL, it does not inherit the advantages of MCE IRL and thus can be inappropriate for stochastic domains. However, similarly to MaxEnt IRL and MCE IRL, it allows for stochastic agent behavior. A disadvantage of REIRL is its requirement for non-expert demonstrations from an arbitrary policy. Klein et al. (2012) reformulate the IRL problem as a structured classification of actions given state- and action-dependent feature counts that are estimated from the demonstrations (SCIRL). Missing feature counts are obtained by querying additional samples of non-optimal actions or by introducing heuristics. However, SCIRL can only be applied if the agent has been following a deterministic policy. Instead, the MDCE-SERD approach in this paper can train models from suboptimal demonstrations. In addition to reward learning in IRL, it optimizes the transition model. A disadvantage of MDCE-SERD is that it needs to repeatedly solve the forward problem and the soft Q-iteration. However, learning better reward functions and transition models is especially beneficial if both need to be transferred to new environments, where it is not possible to query new demonstrations. In such cases, more accurate models will result in better policies.

## 6 Evaluation

We evaluated the MDCE-SERD approach in a grid world navigation task based on satellite images with differing stochastic motion dynamics in forest and open terrain. Furthermore, we tested its generalization capabilities in a transfer learning task. Therefore, we transferred the estimated transition model and the reward function to another satellite image and comparred the resulting policy to the true one. In the learning part of the evaluation, the aim of an agent should be learned solely from demonstrations of a navigation task in a stochastic environment. Figure 1 illustrates the settings of the tasks. In each state the agent can choose from five different actions, which are moving in one of four directions (north, east, south, or west) or staying in the state. The set of successor states is restricted to the current state and the four neighboring states.

On the open terrain (Fig. 1 (c): depicted in light gray) the agent successfully executes a motion action with probability 0.8 and falls with 0.1 to either right or left of the desired direction. The agent's dynamics in the forest (Fig. 1 (c): depicted in dark gray) are more stochastic. Successful transitions occur with probabil-

Michael Herman, Tobias Gindele, Jörg Wagner, Felix Schmitt, Wolfram Burgard

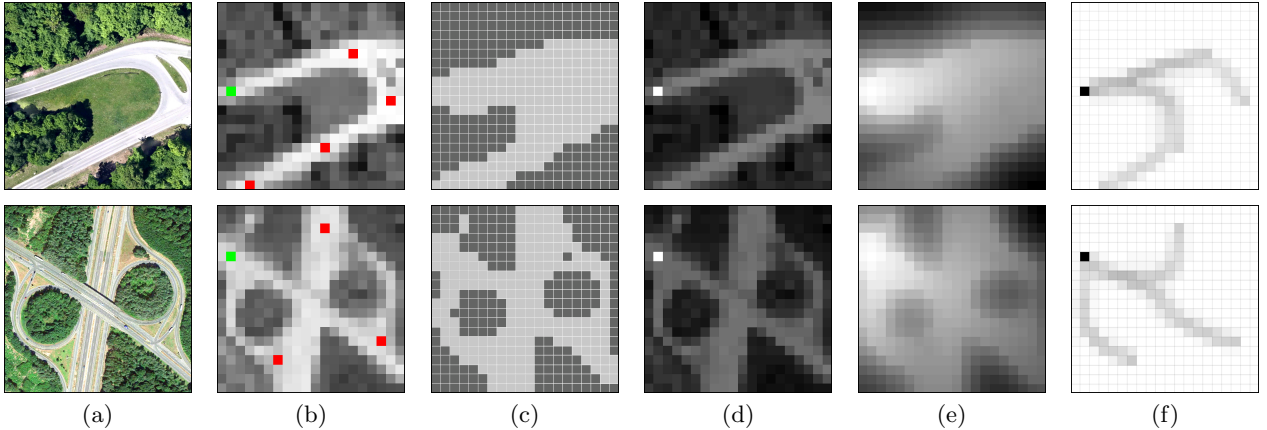

(a)    (b)    (c)    (d)    (e)    (f)

Figure 1: The first row represents the test task and the second row the transfer task. (a) Environment, Map data: Google. (b) Discretized state space. The goal state is indicated in green and start states in red. (c) Forest states are indicated in a dark-gray color and open terrain in light gray. Furthermore, plot (d) shows the reward, (e) the resulting value function, and (f) the expected state frequency.

ity 0.3, otherwise the agent randomly falls into one of the remaining successor states. As a consequence the agent has to trade off short cuts through the forest against longer paths on the open terrain that are more likely to be successful. Staying in a state is always successful. The reward function is a linear combination of two state-dependent features. One of them is the gray scale value of the image, which has been normalized to $[0, 1]$, the other is a goal identity, being 1 in the goal state and 0 otherwise. The feature weights of the true model were set to $\boldsymbol{\theta}_R = (6, 6)^\intercal$ and the discount to 0.99. Fig. 1 (d) illustrates the resulting reward function, which favors roads and especially the goal state.

We computed the stochastic policy of this MDP according to Eq. (5). Then, we sampled trajectories from the resulting policy, which were used as training samples for the evaluation of the learning task. Since the MDCE policy is stochastic, suboptimal expert behavior is considered. For learning, we assumed that the expert has complete knowledge about the true dynamics, so that $\boldsymbol{\theta}_{T_A}$ and $\boldsymbol{\theta}_T$ are equal. This makes it possible to use both terms of Eq. (13) to optimize the transition model parameters. We estimate independent transition models for each action (north, east, south, and west) both in the forest and the open terrain, as well as a shared model for staying. Therefore, 9 models are trained with 5 possible outcomes, resulting in 45 transition model parameters that are energies of Boltzmann distributions. We used an m-estimator with a uniform prior to estimate independent dynamics for each action from the observed transitions. Then, models have been trained from the demonstrations based on random feature weight initializations ($\forall i : \theta_i \in [-10, 10]$) with MDCE IRL, REIRL, and MDCE-SERD for various sizes of demonstration sets. A requirement of REIRL are samples from a arbitrary known policy, such that the IRL problem can be solved with importance sampling. Since only expert demonstrations are available, these samples are generated based on the dynamics estimate and an m-estimated policy from the experts demonstrations. This results in meaningful trajectories and more accurate REIRL estimates.

Figure 2 summarizes the results of the evaluation for varying numbers of expert demonstrations. The first figure illustrates the average log likelihood of demonstrations from the true model on the learned ones, where MDCE-SERD outperforms the other approaches and needs much fewer demonstrations to obtain good estimates. The increase in performance of MDCE-SERD over MDCE IRL is explained by the fact that it can adjust wrongly estimated transition models. It is interesting to note that REIRL performed worse than MDCE IRL. This is probably caused by the fact that REIRL is based on MaxEnt IRL (Ziebart et al., 2008), which doesn't consider the transitition stochasticity to the full extent. This can falsify the learning, especially, if the stochasticity influences the agent's behavior. Figure 2 (b) shows the average Kullback-Leibler divergence between the estimated transition model and the real one. The transition models of REIRL and MDCE IRL have been derived by an m-estimator from the given demonstrations and therefore perform similarly. The transition model of MDCE-SERD has been further optimized simultaneously with the rewards, resulting in more accurate transition models. Figure 2 (c) illustrates the Kullback-Leibler divergence of the estimated policy, where MDCE-SERD outperforms both MDCE IRL and REIRL.

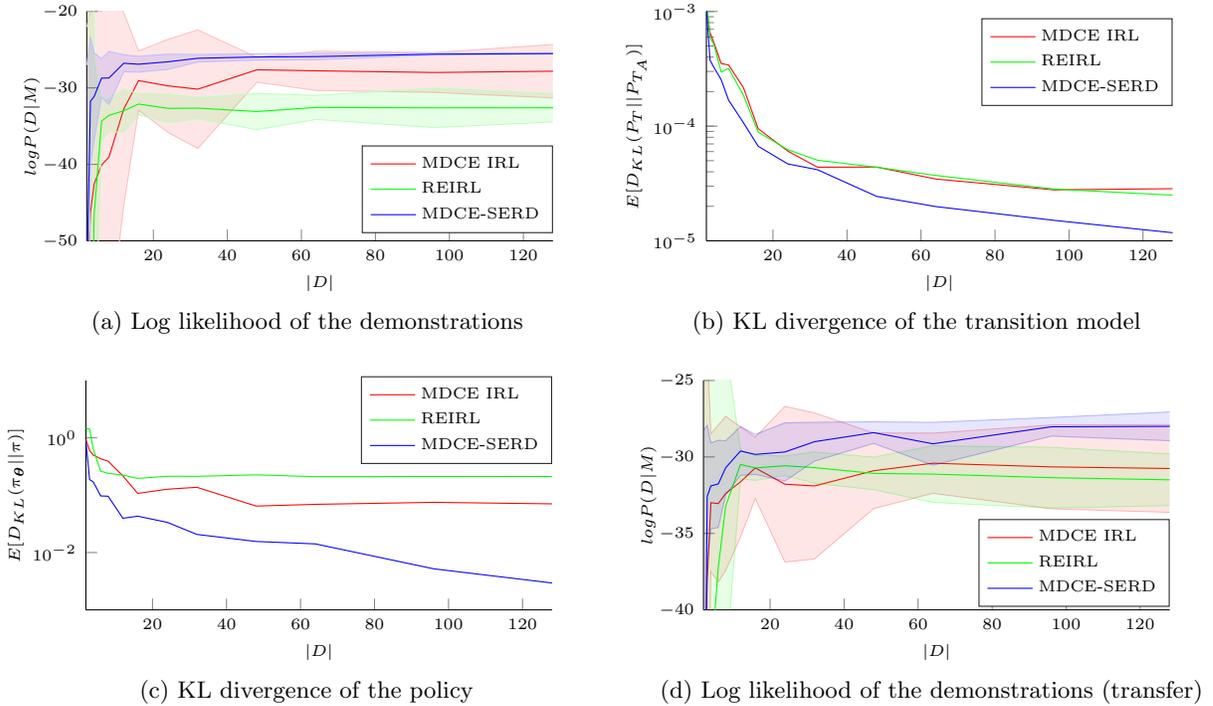

Figure 2: (a) Average log likelihood of demonstrations drawn from the true model under the estimated model. (b) Average Kullback-Leibler divergence between the estimated dynamics and the true ones. (c) Average Kullback-Leibler divergence between the trained stochastic policy and the true one. (d) Average log likelihood of demonstrations drawn from the true model under the estimated model in the transfer task environment.

Then, the estimated models from the learning task have been tranferred to the transfer task environment, where a policy has been computed based on the learned reward function and the estimated transition model. Figure 2 (d) shows the average log likelihood of demonstrations from the true model under the estimated one. MDCE SERD shows an improved performance against the other approaches, probably because it could more accurately estimate the model and therefore generates better policies. Therefore, it can be concluded that if the transition model and the reward function are transferred to a new environment, where the agent suddenly acts in states and actions that have never or rarely been observed, more accurate models can help to generate meaningful policies.

## 7 Conclusion

In this paper we investigated the new problem class of IRL, where both the reward function and the system's dynamics are unknown and need to be estimated from demonstrations. We presented a gradient-based solution, which simultaneously estimates parameters of the transition model and the reward function by taking into account the bias of the demonstrations. To the best of our knowledge, this has not been considered previously and is not possible with current approaches. The evaluation shows that the combined approach estimates models more accurately than MDCE IRL or REIRL, especially in the case of limited data. This is especially beneficial if both the reward function and the transition model are transferred to new environments, since the optimal policy could result in high frequencies of states and actions that were never or rarely observed. In addition, the estimated transition model can be of interest on its own. Future work could extend SERD to partially observable domains or continuous state and action spaces. Furthermore, prior information about rewards or dynamics can be easily introduced, by estimating only a subset of parameters. This allows solving subproblems such as estimating the dynamics for given rewards. An aspect of our approach that can be further exploited is the discrimination between the true system's dynamics and the expert's estimate of it. Examining the relationship between the two could be used to further improve the estimates and even yield policies that exceed the expert's performance.

# Inverse Reinforcement Learning with Simultaneous Estimation of Rewards and Dynamics - Supplementary Material

Michael Herman*†    Tobias Gindele*    Jörg Wagner*    Felix Schmitt*    Wolfram Burgard†

*Robert Bosch GmbH  
D-70442 Stuttgart, Germany

†University of Freiburg  
D-79110 Freiburg, Germany

## Abstract

This document contains supplementary material to the paper *Inverse Reinforcement Learning with Simultaneous Estimation of Rewards and Dynamics* with more detailed derivations, additional proofs to lemmata and theorems as well as larger illustrations and plots of the evaluation task.

## 1 Partial Derivative of the Policy

$$\frac{\partial}{\partial \theta_i} \log \pi_{\boldsymbol{\theta}}(s,a)$$
$$= \frac{\partial}{\partial \theta_i}(Q_{\boldsymbol{\theta}}(s,a) - V_{\boldsymbol{\theta}}(s))$$
$$= \frac{\partial}{\partial \theta_i}\left(Q_{\boldsymbol{\theta}}(s,a) - \log \sum_{a' \in A} \exp(Q_{\boldsymbol{\theta}}(s,a'))\right)$$
$$= \frac{\partial}{\partial \theta_i} Q_{\boldsymbol{\theta}}(s,a) - \frac{\partial}{\partial \theta_i} \log \sum_{a' \in A} \exp(Q_{\boldsymbol{\theta}}(s,a'))$$
$$= \frac{\partial}{\partial \theta_i} Q_{\boldsymbol{\theta}}(s,a)$$
$$\quad - \frac{\sum_{a' \in A}\left[\exp(Q_{\boldsymbol{\theta}}(s,a')) \frac{\partial}{\partial \theta_i} Q_{\boldsymbol{\theta}}(s,a')\right]}{\sum_{a' \in A} \exp(Q_{\boldsymbol{\theta}}(s,a'))}$$
$$= \frac{\partial}{\partial \theta_i} Q_{\boldsymbol{\theta}}(s,a) - \mathbb{E}_{\pi_{\boldsymbol{\theta}}(s,a')}\left[\frac{\partial}{\partial \theta_i} Q_{\boldsymbol{\theta}}(s,a')\right].$$

## 2 Partial Derivative of the Soft Q-Function with Respect to the Individual Parameter Types

The partial derivative can be further simplified if it is taken with respect to the three individual parameter types: the feature weights, the agent's dynamics parameters, and the parameters of the true environment's dynamics:

$$\forall \theta_i \in \boldsymbol{\theta}_R : \frac{\partial}{\partial \theta_i} Q_{\boldsymbol{\theta}}(s,a)$$
$$= f_i(s,a)$$
$$\quad + \gamma \sum_{s' \in S}\left\{P_{\boldsymbol{\theta}_{T_A}}(s'|s,a)\, \mathbb{E}_{\pi_{\boldsymbol{\theta}}(s',a')}\left[\frac{\partial}{\partial \theta_i} Q_{\boldsymbol{\theta}}(s',a')\right]\right\}$$

$$\forall \theta_i \in \boldsymbol{\theta}_{T_A} : \frac{\partial}{\partial \theta_i} Q_{\boldsymbol{\theta}}(s,a)$$
$$= \gamma \sum_{s' \in S}\left[\left(\frac{\partial}{\partial \theta_i} P_{\boldsymbol{\theta}_{T_A}}(s'|s,a)\right) V_{\boldsymbol{\theta}}(s')\right]$$
$$\quad + \gamma \sum_{s' \in S}\left[P_{\boldsymbol{\theta}_{T_A}}(s'|s,a)\, \mathbb{E}_{\pi_{\boldsymbol{\theta}}(s',a')}\left[\frac{\partial}{\partial \theta_i} Q_{\boldsymbol{\theta}}(s',a')\right]\right]$$

$$\forall \theta_i \in \boldsymbol{\theta}_T : \frac{\partial}{\partial \theta_i} Q_{\boldsymbol{\theta}}(s,a) = 0$$

## 3 Proof: Soft Q-iteration is a Contraction Mapping

It has to be shown that the soft Q-iteration is a fixed point iteration with only one fixed point, since this is a requirement of our algorithm. Bloem et al. have shown in Bloem and Bambos (2014) that the soft value iteration operator is a contraction mapping. It has to be proven that the same holds for the soft Q-iteration operator $T_{\boldsymbol{\theta}}^{soft}(\boldsymbol{Q})$. Therefore, we adjust their proof to be valid for the Q-iteration.

The soft Q-iteration operator is defined as

$$T_{\boldsymbol{\theta}}^{soft}(\boldsymbol{Q})(s,a) = \boldsymbol{\theta}_R^\intercal \boldsymbol{f}(s,a) \\ + \gamma \sum_{s' \in S} \left[ P_{\boldsymbol{\theta}_{T_A}}(s'|s,a) \operatorname*{softmax}_{a' \in A}(Q(s',a')) \right]$$

with the function

$$\operatorname*{softmax}_{x_i \in \boldsymbol{x}}(x_i) = \log\left(\sum_{i=1}^{N} \exp(x_i)\right).$$

We will begin with deriving proofs for necessary auxiliary definitions and lemmata. In order to argue about the monotonicity of multidimensional functions, a partial order on $\mathbb{R}^{A \times B}$ is introduced. Then, a property of the softmax function is derived and afterwards the monotonicity of the operator $T_{\boldsymbol{\theta}}^{soft}(\boldsymbol{Q})(s,a) : \mathbb{R}^{|S| \times |A|} \to \mathbb{R}^{|S| \times |A|}$ with respect to the introduced partial order is proven.

**Definition 3.1.** *For $\boldsymbol{x}, \boldsymbol{y} \in \mathbb{R}^{A \times B}$ with $A, B \in \mathbb{N}^+$, the partial order $\preceq$ is defined as $\boldsymbol{x} \preceq \boldsymbol{y} \Leftrightarrow \forall a \in A, b \in B : x_{a,b} \leq y_{a,b}$.*

**Lemma 3.2.** *The* softmax *function has the property that for any $\boldsymbol{x} \in \mathbb{R}^N$ and $d \in \mathbb{R}$ it holds that $\operatorname*{softmax}_{x_i \in \boldsymbol{x}}(x_i + d) = \operatorname*{softmax}_{x_i \in \boldsymbol{x}}(x_i) + d$.*

*Proof.* The property can be easily shown by extracting the variable $d$ from the softmax formulation:

$$\operatorname*{softmax}_{x_i \in \boldsymbol{x}}(x_i + d) = \log\left(\sum_{i=1}^{N} \exp(x_i + d)\right)$$
$$= \log\left(\exp(d) \sum_{i=1}^{N} \exp(x_i)\right)$$
$$= \log\left(\sum_{i=1}^{N} \exp(x_i)\right) + \log(\exp(d))$$
$$= \log\left(\sum_{i=1}^{N} \exp(x_i)\right) + d$$
$$= \operatorname*{softmax}_{x_i \in \boldsymbol{x}}(x_i) + d$$

□

**Lemma 3.3.** *The soft Q-iteration operator $T_{\boldsymbol{\theta}}^{soft}(\boldsymbol{Q})(s,a)$ is monotone, satisfying $\forall \boldsymbol{Q}_m, \boldsymbol{Q}_n \in \mathbb{R}^{|S| \times |A|} : \boldsymbol{Q}_m \preceq \boldsymbol{Q}_n \to T_{\boldsymbol{\theta}}^{soft}(\boldsymbol{Q}_m) \preceq T_{\boldsymbol{\theta}}^{soft}(\boldsymbol{Q}_n)$.*

*Proof.* The partial derivative of the $T_{\boldsymbol{\theta}}^{soft}(\boldsymbol{Q})(s,a)$ with respect to a single value $Q(s_i, a_i)$ is

$$\frac{\partial}{\partial Q(s_i, a_i)} T_{\boldsymbol{\theta}}^{soft}(\boldsymbol{Q})(s,a)$$
$$= \gamma P_{\boldsymbol{\theta}_{T_A}}(s_i|s,a) \frac{\exp(Q(s_i, a_i))}{\sum_{a_j \in A} \exp(Q(s_i, a_j))}.$$

From the definition of the MDP it follows that $\gamma \in [0,1)$ and the probability distribution $P_{\boldsymbol{\theta}_{T_A}}(s_i|s,a) \in [0,1]$. As $\forall x_i \in \mathbb{R} : \exp(x_i) \in (0, +\infty)$, all terms of the partial derivative $\frac{\partial}{\partial Q(s_i, a_i)} T_{\boldsymbol{\theta}}^{soft}(\boldsymbol{Q})(s,a)$ are positive or zero, which finishes the proof that $\frac{\partial}{\partial Q(s_i, a_i)} T_{\boldsymbol{\theta}}^{soft}(\boldsymbol{Q})(s,a) \geq 0$. □

Based on Lemma 3.2 and 3.3 it is possible to derive the proof that the soft Q-iteration is a contraction mapping. We transfer the proof of Bloem et al. Bloem and Bambos (2014) for the value iteration and adjust it, such that it applies for the Q-iteration.

**Theorem 3.4.** *The soft Q-iteration operator $T_{\boldsymbol{\theta}}^{soft}(\boldsymbol{Q})(s,a)$ is a contraction mapping with only one fixed point. Therefore, it is Lipschitz continuous $\|T_{\boldsymbol{\theta}}^{soft}(\boldsymbol{Q}_m) - T_{\boldsymbol{\theta}}^{soft}(\boldsymbol{Q}_n)\|_\infty \leq L \|\boldsymbol{Q}_m - \boldsymbol{Q}_n\|_\infty$ for all $\boldsymbol{Q}_m, \boldsymbol{Q}_n \in \mathbb{R}^{|S| \times |A|}$ with a Lipschitz constant $L \in [0, 1)$.*

*Proof.* Consider $\boldsymbol{Q}_m, \boldsymbol{Q}_n \in \mathbb{R}^{|S| \times |A|}$. There exists a distance $d$ under the supremum norm, for which $\exists d \in \mathbb{R}_0^+ : \|\boldsymbol{Q}_m - \boldsymbol{Q}_n\|_\infty = d$ holds and therefore

$$-d\mathbf{1} \preceq \boldsymbol{Q}_m - \boldsymbol{Q}_n \preceq d\mathbf{1}$$

with $\mathbf{1} = (1)_{k,l}$, where $1 \leq k \leq |S|, 1 \leq l \leq |A|$. Since $d$ bounds the components of the vector difference $\boldsymbol{Q}_m - \boldsymbol{Q}_n$, it can be derived that $\boldsymbol{Q}_m \preceq \boldsymbol{Q}_n + d\mathbf{1}$ and $\boldsymbol{Q}_n \preceq \boldsymbol{Q}_m + d\mathbf{1}$. For both cases, the monotonicity condition of Lemma 3.3 is satisfied, which allows for the following inequality: $T_{\boldsymbol{\theta}}^{soft}(\boldsymbol{Q}_m) \preceq T_{\boldsymbol{\theta}}^{soft}(\boldsymbol{Q}_n + d\mathbf{1})$. By applying Lemma 3.2, it follows that $\forall s \in S, a \in A$

$$T_{\boldsymbol{\theta}}^{soft}(\boldsymbol{Q}_m)(s,a)$$
$$\leq T_{\boldsymbol{\theta}}^{soft}(\boldsymbol{Q}_n + d\mathbf{1})(s,a)$$
$$= \boldsymbol{\theta}_R^\intercal \boldsymbol{f}(s,a)$$
$$+ \gamma \sum_{s' \in S} \left[ P_{\boldsymbol{\theta}_{T_A}}(s'|s,a) \operatorname*{softmax}_{a' \in A}(Q(s',a') + d) \right]$$
$$= \boldsymbol{\theta}_R^\intercal \boldsymbol{f}(s,a)$$
$$+ \gamma \sum_{s' \in S} \left[ P_{\boldsymbol{\theta}_{T_A}}(s'|s,a) \operatorname*{softmax}_{a' \in A}(Q(s',a')) + d \right]$$
$$= \boldsymbol{\theta}_R^\intercal \boldsymbol{f}(s,a)$$
$$+ \gamma \sum_{s' \in S} \left[ P_{\boldsymbol{\theta}_{T_A}}(s'|s,a) \operatorname*{softmax}_{a' \in A}(Q(s',a')) \right] + \gamma d$$
$$= T_{\boldsymbol{\theta}}^{soft}(\boldsymbol{Q}_n)(s,a) + \gamma d$$

In vector notation this results in $T_{\boldsymbol{\theta}}^{soft}(\boldsymbol{Q}_m) \preceq T_{\boldsymbol{\theta}}^{soft}(\boldsymbol{Q}_n) + \gamma d\mathbf{1}$. As from the symmetric definition of $-d\mathbf{1} \preceq \boldsymbol{Q}_m - \boldsymbol{Q}_n \preceq d\mathbf{1}$, it has been derived that $\boldsymbol{Q}_n \preceq \boldsymbol{Q}_m + d$, it consequently follows that $T_{\boldsymbol{\theta}}^{soft}(\boldsymbol{Q}_n) \preceq T_{\boldsymbol{\theta}}^{soft}(\boldsymbol{Q}_m) + \gamma d\mathbf{1}$. To finish the proof, it has to be shown that the soft Q-iteration operator is Lipschitz continuous with $L \in [0, 1)$. This can be done by combining the related inequations of the operator:

$$-\gamma d\mathbf{1} \preceq T_{\boldsymbol{\theta}}^{soft}(\boldsymbol{Q}_m) - T_{\boldsymbol{\theta}}^{soft}(\boldsymbol{Q}_n) \preceq \gamma d\mathbf{1}$$
$$||T_{\boldsymbol{\theta}}^{soft}(\boldsymbol{Q}_m) - T_{\boldsymbol{\theta}}^{soft}(\boldsymbol{Q}_n)||_\infty \leq \gamma d$$
$$||T_{\boldsymbol{\theta}}^{soft}(\boldsymbol{Q}_m) - T_{\boldsymbol{\theta}}^{soft}(\boldsymbol{Q}_n)||_\infty \leq \gamma ||\boldsymbol{Q}_m - \boldsymbol{Q}_n||_\infty$$

This proves that the soft Q-iteration operator $T_{\boldsymbol{\theta}}^{soft}(\boldsymbol{Q})$ is Lipschitz continuous with a Lipschitz constant $L = \gamma$ and $\gamma \in [0, 1)$, resulting in a contraction mapping. As this holds for the whole input space of $\mathbb{R}^{|S| \times |A|}$, two points would always contract, so there cannot exist two fixed points. □

## 4 Proof: The Converged Soft Q-Function is Differentiable

**Theorem 4.1.** *The converged soft Q-function is differentiable with respect to $\boldsymbol{\theta}$.*

*Proof.* Since we provide an iterative formula for the gradient of the converged soft Q-function $\tilde{Q}(s, a)$, we need to revisit the soft Q-iteration operator $T_{\boldsymbol{\theta}}^{soft}(\boldsymbol{Q}) : \mathbb{R}^{S \times A} \mapsto \mathbb{R}^{S \times A}$, element-wise defined as

$$T_{\boldsymbol{\theta}}^{soft}(Q(s', a'))[s, a] = \boldsymbol{\theta}_R^\intercal \boldsymbol{f}(s, a) + \gamma \sum_{s' \in S} \left[ P_{\boldsymbol{\theta}_{T_A}}(s'|s, a) \log(\sum_{a' \in A} \exp(Q(s', a'))) \right].$$

It is ensured that repeatedly applying $T_{\boldsymbol{\theta}}^{soft}(\boldsymbol{Q})$ to an initial $\boldsymbol{Q}_0$ converges to a fixed-point $\tilde{\boldsymbol{Q}}$ given $\gamma \in [0, 1)$, as the soft Q-operator converges [see Section 3]. $T_{\boldsymbol{\theta}}^{soft}(\boldsymbol{Q})$ is differentiable with respect to both $\boldsymbol{Q}$ and $\boldsymbol{\theta}$ as being the composition of differentiable functions. This requires the transition model $P_{\boldsymbol{\theta}_{T_A}}$ to be differentiable with respect to $\boldsymbol{\theta}$, too. We now apply the implicit function theorem Krantz and Parks (2002) to compute the derivative $\frac{\partial}{\partial \boldsymbol{\theta}} \tilde{\boldsymbol{Q}}_{\boldsymbol{\theta}}$ given by the equation

$$T_{\boldsymbol{\theta}}^{soft}(\boldsymbol{Q})(s, a) - Q(s, a) = 0.$$

The theorem states that if the Jacobian $\frac{\partial}{\partial \boldsymbol{Q}}[T_{\boldsymbol{\theta}}^{soft}(\boldsymbol{Q}) - \boldsymbol{Q}]$ is invertible at $\tilde{\boldsymbol{Q}}_{\boldsymbol{\theta}}$, the derivative $\frac{\partial}{\partial \boldsymbol{\theta}} \tilde{\boldsymbol{Q}}_{\boldsymbol{\theta}}$ exists and is given by

$$\frac{\partial}{\partial \boldsymbol{\theta}} \tilde{\boldsymbol{Q}}_{\boldsymbol{\theta}} = \left( \frac{\partial}{\partial \boldsymbol{Q}} [T_{\boldsymbol{\theta}}^{soft}(.) - .] \right)^{-1} \frac{\partial}{\partial \boldsymbol{\theta}} T_{\boldsymbol{\theta}}^{soft}(.) \, (\tilde{\boldsymbol{Q}}_{\boldsymbol{\theta}}).$$

Since the partial derivative of the operator $T_{\boldsymbol{\theta}}^{soft}(\boldsymbol{Q})$ has already been derived in Section 3, the Jacobian of $T_{\boldsymbol{\theta}}^{soft}(\boldsymbol{Q}) - \boldsymbol{Q}$ is

$$\frac{\partial}{\partial \boldsymbol{Q}}[T_{\boldsymbol{\theta}}^{soft}(\tilde{\boldsymbol{Q}}_{\boldsymbol{\theta}}) - \tilde{\boldsymbol{Q}}_{\boldsymbol{\theta}}]([s, a], [s', a'])$$
$$= -\delta(a' = a, s' = s) + \gamma P_{\boldsymbol{\theta}_{T_A}}(s'|s, a)$$
$$\cdot \frac{1}{\sum_{a'' \in A} \exp(Q(s', a''))} \exp(Q(s', a'))$$
$$= -\delta(a' = a, s' = s) + \gamma \underbrace{\pi(a'|s') P_{\boldsymbol{\theta}_{T_A}}(s'|a, s)}_{M_{[s,a],[s',a']}}.$$

It holds $1 > \gamma \geq ||\gamma \boldsymbol{M}||_\infty$ in the $L\infty$ induced matrix-norm defined as $||A||_\infty := \max_x \frac{|Ax|_\infty}{|x|_\infty} = \max_i \sum_j |A_{i,j}|$, as

$$\max_{[s,a]} \sum_{[s',s']} |\gamma M_{[s,a],[s',a']}|$$
$$= \max_{[s,a]} \sum_{[s',a']} |\gamma \pi(a'|s') P_\theta(s'|a, s)| = \gamma.$$

Hence, $(\gamma \boldsymbol{M} - I)^{-1}$ exists and is given by the *Neuman operator-series* $-\sum_{i=0}^\infty (\gamma \boldsymbol{M})^i$. Since the Jacobian is invertible it is proven that the partial derivative of the converged soft Q-function $\frac{\partial}{\partial \boldsymbol{\theta}} \tilde{\boldsymbol{Q}}_{\boldsymbol{\theta}}$ with respect to the parameters $\boldsymbol{\theta}$ exists. □

## 5 Grid World Terrain Motion Task

This section provides larger illustrations and results for the evaluation task. Figure 1 and 2 illustrate the environment of the training and transfer task. The results are summarized in Figure 3, where (a), (b), (c) are results on the training task and (d) presents the performance on the transfer task. We used Welch's t-test Welch (1947) to verify that the differences of the mean log likelihood of the demonstrations under the trained models in Figure 3 (a) and (d) are statistically significant ($p < 0.05$). In the training task, the performance of SERD against the other approaches is statistically significant for sample set sizes that are larger than 3, while in the transfer task statistical significance is given at least for demonstration set sizes larger than 12.

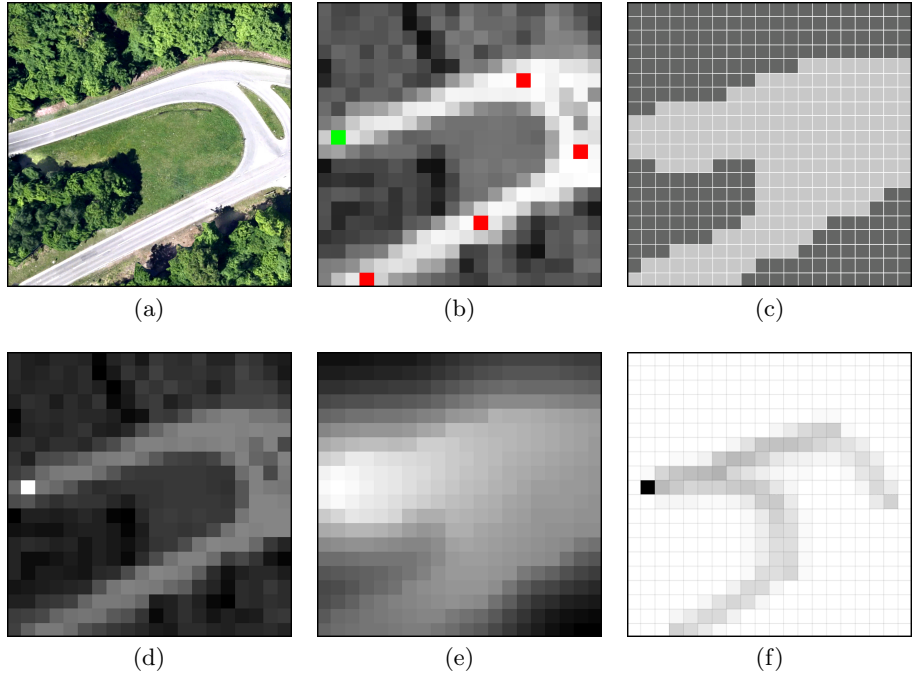

Figure 1: The training and test task. (a) Environment, Map data: Google. (b) Discretized state space. The goal state is indicated in green and start states in red. (c) Forest states are indicated in a dark-gray color and open terrain in light gray. Furthermore, plot (d) shows the reward, (e) the resulting value function, and (f) the expected state frequency.

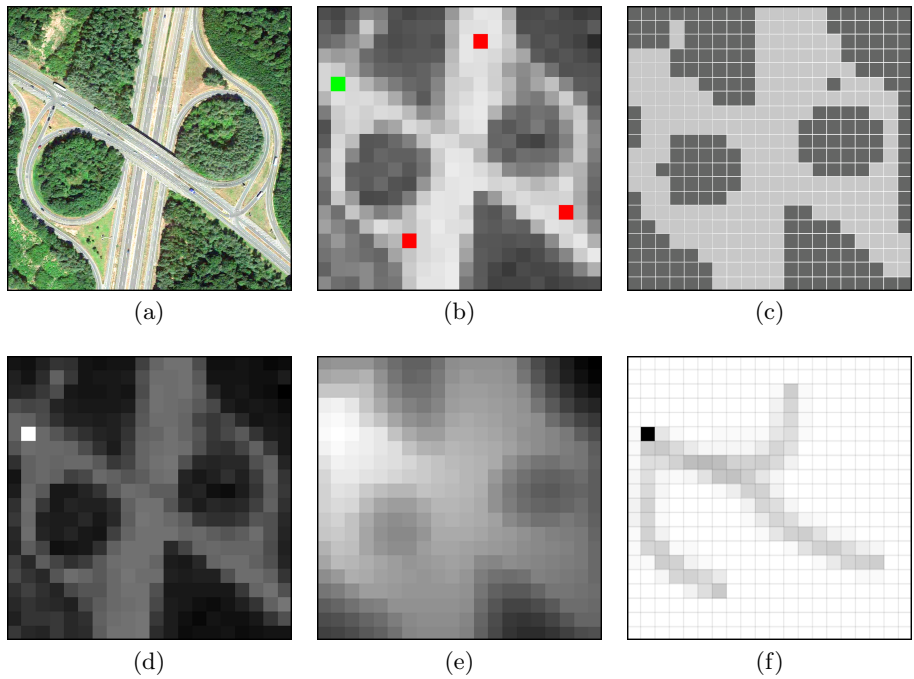

Figure 2: The transfer task. (a) Environment, Map data: Google. (b) Discretized state space. The goal state is indicated in green and start states in red. (c) Forest states are indicated in a dark-gray color and open terrain in light gray. Furthermore, plot (d) shows the reward, (e) the resulting value function, and (f) the expected state frequency.

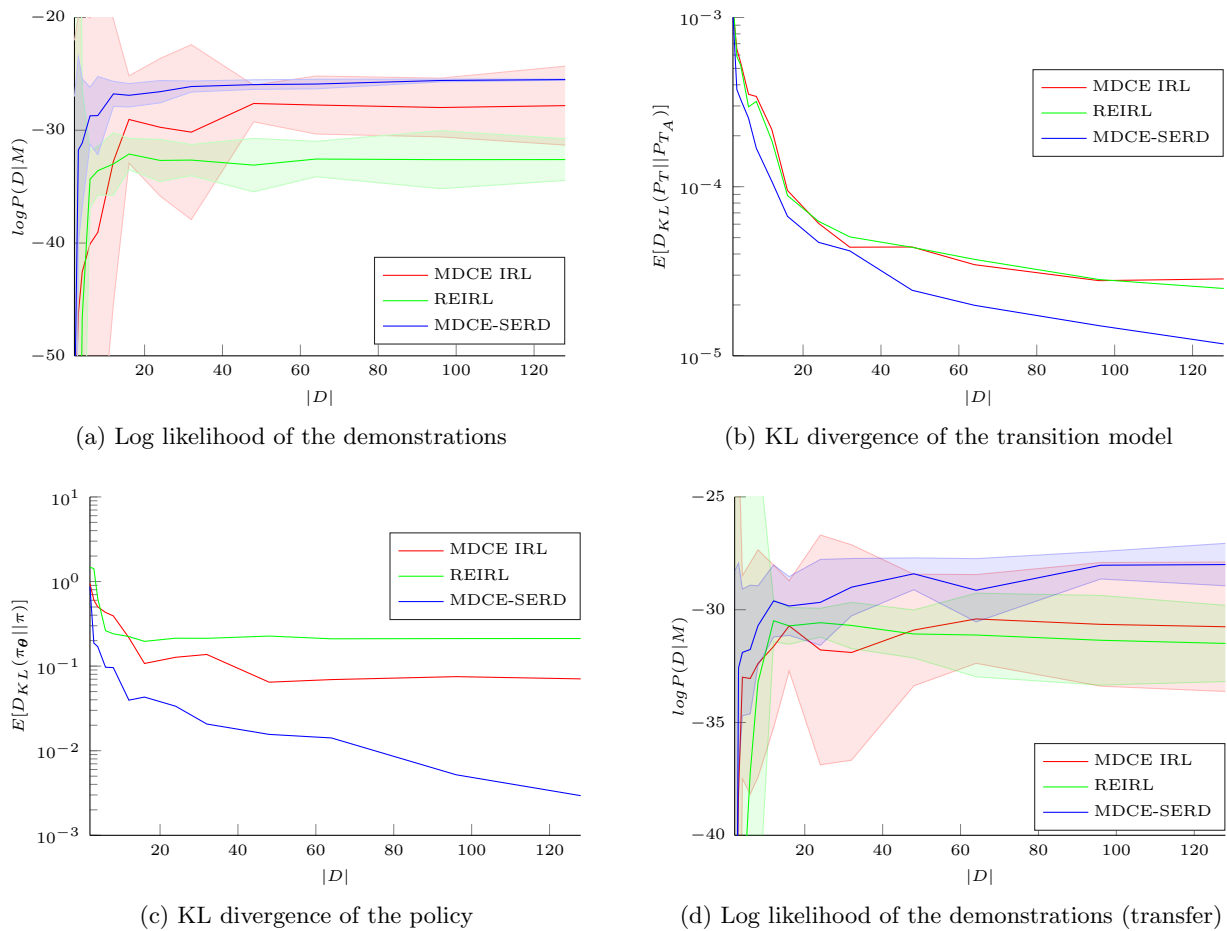

Figure 3: (a) Average log likelihood of demonstrations drawn from the true model under the estimated model. (b) Average Kullback-Leibler divergence between the estimated dynamics and the true ones. (c) Average Kullback-Leibler divergence between the trained stochastic policy and the true one. (d) Average log likelihood of demonstrations drawn from the true model under the estimated model in the transfer task environment.